\documentclass[journal]{IEEEtai}

\usepackage[colorlinks,urlcolor=blue,linkcolor=blue,citecolor=blue]{hyperref}

\usepackage{color,array}
\pdfoutput=1

\usepackage{graphicx}  
\usepackage{multirow}
\usepackage{amssymb}  
\usepackage{amsmath}
\usepackage{booktabs}  
\usepackage{cite}
\usepackage{epsfig}
\usepackage{subcaption}

\begin{document}
	
\title{A Fast Attention Network for Joint Intent Detection and Slot Filling on Edge Devices}

\author{Liang~Huang,~\IEEEmembership{Member,~IEEE,}
		Senjie~Liang,
		Feiyang~Ye,
		and Nan~Gao
\thanks{L.~Huang and N.~Gao are with the College of Computer Science and Technology, Zhejiang University of Technology, Hangzhou, China 310058, (e-mail: \{lianghuang, gaonan\}@zjut.edu.cn).}
\thanks{S.~Liang and F.~Ye are with the College of Information Engineering, Zhejiang University of Technology, Hangzhou, China 310058,  (e-mail: \{senjieliang, feiyangye\}@zjut.edu.cn).}}	


\maketitle
	
	\begin{abstract}
Intent detection and slot filling are two main tasks in natural language understanding and play an essential role in task-oriented dialogue systems. The joint learning of both tasks can improve inference accuracy and is popular in recent works. However, most joint models ignore the inference latency and cannot meet the need to deploy dialogue systems at the edge. In this paper, we propose a Fast Attention Network (FAN) for joint intent detection and slot filling tasks, guaranteeing both accuracy and latency. Specifically, we introduce a clean and parameter-refined attention module to enhance the information exchange between intent and slot, improving semantic accuracy by more than 2\%. FAN can be implemented on different encoders and delivers more accurate models at every speed level. Our experiments on the Jetson Nano platform show that FAN inferences fifteen utterances per second with a small accuracy drop, showing its effectiveness and efficiency on edge devices.
	\end{abstract} 

	\begin{IEEEImpStatement}
       Dialogue systems at the edge are an emerging technology in real-time interactive applications. They improve the user experience with low latency and secure privacy without transferring personal data to the cloud servers. However, it is challenging to guarantee inference accuracy and low latency on hardware-constrained devices with limited computation, memory storage, and energy resources. The neural network models we introduce in this paper overcome these limitations. With a significant increase in semantic accuracy by more than 2\% after adopting our algorithms, the technology reduces the inference latency to less than 100ms. From this viewpoint, our approaches accelerate the boosting of secure personal assistants to end-users.
\end{IEEEImpStatement} 


\section{Introduction}
\IEEEPARstart{D}{ialogue} systems at the edge have tremendous potential to power secure and real-time interactive applications \cite{bertino2020artificial}, i.e., augmented/virtual reality, autonomous vehicles, and robots. They store personal data at the edge to get void of privacy leakage, as opposed to the cloud-based commercial services, i.e., Google Assistant and Amazon Alexia. Dialogue systems such as task-oriented dialogue systems and intelligent personal assistants have been deployed near the user at different edge platforms, i.e., Raspberry Pi \cite{xu2020cha}, Jetson Nano \cite{pandelea2021emotion}, and smartphones. The major challenge is guaranteeing real-time user experience on hardware-constrained devices with limited computation, memory storage, and energy resources.

Natural language understanding (NLU) \cite{young2013pomdp} is crucial for understanding user input for effective human-computer interaction to establish an innovative and efficient human-machine dialogue system. NLU generally includes both intent detection and slot filling \cite{tur2011spoken}. Intent detection focuses on automatically identifying the intent of user utterances, which can be considered a classification problem. Slot filling extracts semantic constituents from the natural language utterances to provide essential information for the system to take the following action, which can be considered a sequence label problem. Intent detection and slot filling were performed separately in earlier studies in a pipeline approach \cite{jeong2008triangular}, which first classifies the intent of an utterance and then uses the extra intent information to aid slot filling. Commonly used approaches for intent detection are support vector machine (SVM) \cite{haffner2003optimizing} and recurrent neural network (RNN) \cite{yao2014spoken}, and for slot filling are conditional random field (CRF) \cite{raymond2007generative} and RNN. However, an incorrect intent prediction will possibly mislead the successive slot filling in the pipeline approaches. 
	
	The trend is to develop a joint model for both intent detection and slot filling tasks to avoid error propagation in the pipeline approaches. Some joint models apply a joint loss function to connect the two tasks \cite{chen2016syntax, zhang2016joint, hakkani2016multi, liu2016attention}. Some models take advantage of the close relationship between two tasks and use some structures, i.e., Slot-gate \cite{goo2018slot},  Stack-Propagation \cite{qin-etal-2019-stack}, Co-Interactive \cite{qin2021co}, to model the relationship between both tasks explicitly. Modeling the relationship between the two tasks enables these models to achieve significant performance improvements and thus demonstrates the effectiveness of this approach. Recently, pre-trained language models, such as bidirectional encoder representation from transformers (BERT) \cite{devlin-etal-2019-bert}, are widely used in various natural language processing (NLP) tasks. BERT has  contextual solid representation capabilities. Some works \cite{chen2019bert, zhang2019joint} have used it in joint intent detection and slot filling models and gotten a considerable performance boost, bringing the accuracy of joint model predictions to a new level.
	
	However, most of the previous work focuses on improving model prediction accuracy, and a few works consider the inference latency. SlotRefine \cite{wu2020slotrefine} uses a two-pass iteration mechanism to replace CRF for slot decoding while considerably speeding up the decoding. From our perspective, although large pre-trained language models such as BERT bring significant performance gains, these models also have increasing demands on computational resources and memory \cite{khattab2020colbert}. Due to high latency or memory overflow, these models are challenging to deploy directly in certain computing resource-constrained conditions, such as on edge devices \cite{pandelea2021emotion}. Edge computing is a real need due to the response latency and privacy issues associated with computing from the cloud \cite{liu2021bringing}. Therefore, different model compression methods are proposed in the literature to reduce the model parameters and speed up the inference time, i.e., Knowledge Distillation \cite{sanh2019distilbert,jiao2020tinybert,liu2020fastbert,sun2020mobilebert}, Quantization \cite{kim2021bert}, and Structured Pruning \cite{wang2020structured}. To achieve a better balance between speed and accuracy on edge devices, we argue that the following problems should be addressed: 1) Although large-scale pre-training language models such as BERT have greatly improved the inference accuracy, their massive number of parameters makes the inference time expensive; 2) Although prior works such as SlotRefine speed up the decoding, they do not utilize pre-trained knowledge to enhance accuracy; 3) Although Co-Interactive \cite{qin2021co} has achieved significant accuracy gains, their models are complex and have many parameters. Therefore, they incur high latency and memory usage.
	
	In this paper, we propose a \textbf{F}ast \textbf{A}ttention \textbf{N}etwork (FAN) for joint intent detection and slot filling that aims to speed up the model inference without  compromising  the accuracy. FAN is generic for various pre-trained language models. In this paper, we used three different pre-trained language models: BERT, DistilBERT \cite{sanh2019distilbert}, and TinyBERT \cite{jiao2020tinybert}, as an encoder to learn the representation of utterances. In our FAN, We designed a parameter-refined attention module to carry out the two-way information interaction between intentions and slots. The model's performance is improved for the intent detection and slot filling tasks without significantly impacting the model speed and parameters. The attention module consists of a label attention layer and a multi-head self-attention layer, where the label attention layer  integrates the result information of the two tasks into the representation of the utterance. The multi-head self-attention layer bidirectionally shares and exchanges information for intent detection and slot filling to promote each other, instead of only considering the single flow of information from intents to slots as in previous work. In summary, the contributions of this paper are as follows:
	\begin{enumerate}
		\item We propose a novel model framework FAN to jointly model intent detection and slot filling tasks. FAN uses a parameter-refined attention module for information interaction between intent and slot. Numerical experiments show that such a clean scheme achieves state-of-the-art inference accuracy on different datasets.
		\item We implement FAN on various pre-trained language models and experimentally show that FAN delivers more accurate models at every speed level. When TinyBERT is  the encoder, FAN improves the semantic accuracy by more than 2.0\%.
		\item We deploy FAN on popular edge devices. It inferences fifteen utterances per second on the Jecson Nano platform while guaranteeing a comparable accuracy.
	\end{enumerate}

	The rest of our paper is structured as follows. Section ~\ref{sec:relatedwork} reviews the related works on joint model and model compression. Section~\ref{sec:Approach} gives a detailed description of our model. Section~\ref{sec:result} presents experimental results and analysis. Section~\ref{sec:conclusion} summarizes this work and the future direction.

\section{Related Work} \label{sec:relatedwork}
\subsection{Joint Intent Detection and Slot Filling}
Recently, some joint models have overcome the error propagation caused by the pipelined approaches. Goo et al. \cite{goo2018slot} proposed a slot-gated mechanism to pass the intent information into the slot filling task for interactions between intent and slots. Li et al. \cite{li2018self} proposed a new gating mechanism based on self-attention and multi-layer perceptron (MLP) to transfer intent information to the slot filling task. Qin et al. \cite{qin-etal-2019-stack} adopted a Stack-Propagation framework for interactions between intent and slots, which can directly use the intent result information as input for slot filling. E et al. \cite{niu2019novel} proposed an SF-ID network in the middle of the LSTM-based encoder and the CRF-based decoder. The SF subnet applies intent information to the slot filling task, while the ID subnet uses slot information in the intent detection task. Wu et al. Chen et al. \cite{chen2019bert} used BERT as the encoder in the joint model for the first time, bringing the accuracy of the joint model to a new level. \cite{wu2020slotrefine} proposed Two-pass Refine Mechanism to solve the problem of the uncoordinated slots and to speed up model inference by replacing CRF. In the first pass, the model is used to predict the ``B'' label, and in the second pass, the predicted ``B'' label information is sent back to the model to predict the `I' label. Qin et al. \cite{qin2021co} proposed a Co-Interactive Transformer based on the transformer encoder for bidirectional information exchange between intent and slots. Wei et al. \cite{wei2021joint} proposed a wheel-graph structure based on the graph attention network (GAT) to use the correlation between intent and slots. 

\subsection{Model Compression}
Pre-trained language models such as BERT, XLNet \cite{yang2019xlnet}, and RoBERTa \cite{liu2019roberta} were widely used in NLP tasks and achieved significant performance improvements. For example, BERT has powerful semantic representation capabilities and can be used for various downstream tasks through simple fine-tuning. For intent detection and slot filling, the use of BERT has also brought about a significant effect improvement in joint models \cite{chen2019bert,zhang2019joint}. However, although these models have brought significant improvement, these models usually have hundreds of millions of parameters, which consumes a lot of computational resources and experiences a long inference time in practical applications. Therefore, some works tried to compress the pre-trained model. ALBERT \cite{lan2019albert} incorporates embedding factorization and cross-layer parameter sharing to reduce model parameters. Since ALBERT does not reduce the hidden size or layers of the transformer block, it still has a large amount of computation and is time-consuming in the prediction process. DistilBERT \cite{sanh2019distilbert}  performs distillation at the pre-training stage on a large-scale corpus. DistilBERT, which consists of 6 bidirectional transformer encoder layers, is 40\% smaller and 60\% faster than BERT, but it still retains 97\% of the language understanding capability. TinyBERT \cite{jiao2020tinybert} uses a new two-stage learning framework, which performs Transformer distillation at both the pre-training and task-specific learning stages. This two-stage learning method reduces the size of the BERT model by 87\%. The TinyBERT with four transformer encoder layers is 7.5x smaller and 9.4x faster than BERT and still retains more than 96.8\% of the performance of BERT on the GLUE \cite{wang2018glue} benchmark. MobileBERT \cite{sun2020mobilebert} is a thin version of BERT$_{LARGE}$. The knowledge in MobileBERT is transferred from a specially designed teacher model, which is an inverted-bottleneck incorporated BERT$_{LARGE}$ model.

\subsection{Edge Intelligence}
	A computational gap has arisen between deep learning algorithms with high computational demands and edge devices with low computational power. Many high-precision deep learning algorithms cannot be deployed on edge devices. Therefore, many approaches have arisen to address this computational gap \cite{liu2021bringing}. Pandelea et al. \cite{pandelea2021emotion} combined a large transformer as a feature extractor with a simple classifier, deployed it on Jetson Nano and two smartphones, and optimized latency and performance using dimensionality reduction and pre-training. Xu et al. \cite{xu2020cha} proposed an edge-based caching framework for voice assistant systems called CHA on three edge devices, Raspberry Pi, Intel Fog Reference Design, and Jetson AGX Xavier.

\section{Method} \label{sec:Approach}

\subsection{Problem Setup}    
Consider a user's utterance with $T$ tokens, $\{ x_i | i \in \mathcal{T}, \mathcal{T}= \{1,\cdots,T\}\}$, where $ x_{i}$ represents the $i$-th token. The main goal is to jointly predict the utterance's unique intent label $ y^{I} $ and a set of task-specific slot labels $ \{y_{i}^{S} | i \in \mathcal{T}\}$ for all tokens $\{ x_i | i \in \mathcal{T}\}$ on a one-to-one basis. We represent the slot labels in the BIO form  \cite{ramshaw1999text}, denoting whether a token is the beginning of a slot (``B''), inside a slot (``I''), or outside any slot (``O''). For example, the utterance ``find fish story'' from the Snips dataset \cite{coucke2018snips} is labeled the intent, ``Search Screening Event'', and its three tokens, i.e., ``find'', ``fish'', and ``story'' are labeled with different slots, ``O'', ``B-movie\_name'', and ``I-movie\_name", respectively. 

\begin{figure*}
    \centering 
    \includegraphics[width=1.8 \columnwidth]{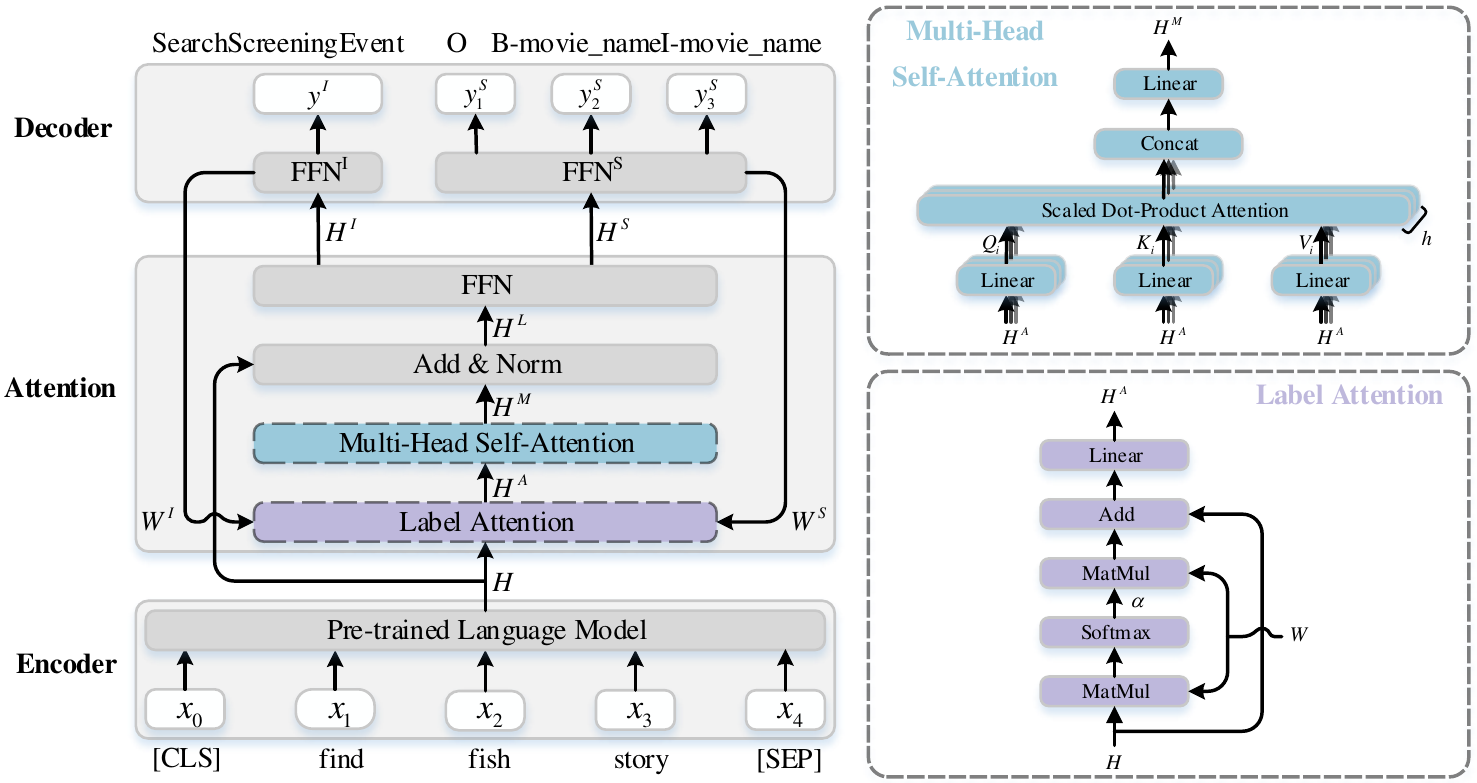}
    \caption{The overview architecture of FAN.}
    \label{fig:model}
\end{figure*}
In the following subsections, we introduce the details of the proposed FAN framework, as illustrated in Fig.~\ref{fig:model}. Specifically, it comprises an encoder module, an attention module, and a decoder module. The encoder module extracts the semantic representation vector $\mathbf{H}$ from the input utterance $\mathbf{x}$. Then, we propose a simple and effective attention module to purify the intent and slot information from $\mathbf{H}$, denoted as $\mathbf{H^I}$ and $\mathbf{H^S}$, respectively. Based on this, the decoder module predicts the corresponding intent label $y^I$ and the slot labels $\{y_i^s\}$.
	
	\subsection{Encoder Module}
	For an utterance $\mathbf{x}$, we insert a special token ([CLS]) at the beginning of the utterance as the first token, denoted as $x_0$ and a special token ([SEP]) at the end of utterance as the final token, denoted as $x_{T}$. Given the input utterance consisting of $T+1$ tokens $ \mathbf{x} = (x_{0},x_{1},\cdots,x_{T}) $ into the pre-trained language model based encoder, and then it produces the semantic representation of tokens, $ \mathbf{H}\in \mathbb{R}^{(T+1)\times d} $. Where $ d $ represents the hidden dimension of the pre-trained language model.

	We choose BERT, DisitlBERT, and TinyBERT as our encoders. Among them, BERT can achieve the best performance. DistilBERT and TinyBERT can significantly improve the model's speed without losing too much accuracy and alleviate the problem of excessive BERT model parameters to a certain extent. DistilBERT and TinyBERT are all knowledge distillation versions of BERT, which have a similar structure to BERT. 
	
	The input representation concatenates the WordPiece embedding \cite{wu2016google}, segment embedding, and position embedding for BERT and TinyBERT. Since DistilBERT does not have NSP pre-training tasks \cite{sanh2019distilbert}, its segment embedding is absent. However, segment embedding has no discrimination for intent detection and slot filling tasks. 
	
	\subsection{Attention Module}
	 As shown in Fig.~\ref{fig:model}, the attention module comprises a label attention layer and a multi-head self-attention layer. We input the semantic representation of tokens $ \mathbf{H} $ into label attention layer and obtain the resulted representation vectors $ \mathbf{H^{A}} $. We feed the output of label attention layer $ \mathbf{H^{A}} $ into the multi-head self-attention layer and obtain the output, $\mathbf{H^M}$. Bypassing $ \mathbf{H^M} $ through a residual connection \cite{he2016deep}, a layer normalization \cite{ba2016layer}, and a two-layer fully connected feed-forward network (FFN). we obtain the output of the attention module, $ \mathbf{H_{}'} $.

	\subsubsection{Label Attention Layer}
	The architecture of the label attention layer is illustrated in Fig.~\ref{fig:model}. We perform the label attention layer to integrate the predicted label information into the representation of tokens $ \mathbf{H} $. 
	
	Specifically, we first obtain the weight of the fully connected feed-forward network in decoder module, which is denoted by $ \mathbf{W^{I}}\in \mathbb{R}^{d\times N^{i}} $, $ N^{i} $ represents the number of intent labels, and $ \mathbf{W^{S}}\in \mathbb{R}^{d\times N^{s}} $, where $ N^{s} $ represents the number of slot label. Then we concatenate the two weights $ \mathbf{W^{I}} $ and $ \mathbf{W^{S}} $ together as $ \mathbf{W}=[\mathbf{W^{I}}, \mathbf{W^{S}}] $.
	
	In practice, we multiply the representation vector $ \mathbf{H} $ with the weight of the decoder $ \mathbf{W} $, and map it to a probability distribution through the softmax layer:
		\begin{equation}
			\boldsymbol{\alpha}=softmax\left(\mathbf{H}\mathbf{W}\right)
		\end{equation}
	where $ \boldsymbol{\alpha} $ is the output of the softmax layer. Then, we integrate the information contained in $ \boldsymbol{\alpha} $ into the representation of tokens:
		\begin{equation}
			\mathbf{H^{A}}=\left(\mathbf{H}+\boldsymbol{\alpha}\left(\mathbf{W^{I}}\right)^{T}\right)\mathbf{W^A}
		\end{equation}
	where $ \mathbf{H^{A}} $ is the representation of tokens after the label attention layer. $\mathbf{W^A}$ is the trainable parameters of linear projectors.
	
	\subsubsection{Multi-Head Self-Attention Layer}
	Inspired by multi-head self-attention in machine translation, we exploit multi-head self-attention to model a bidirectional connection between intent and slots and capture information about the close relationship between two tasks. The architecture of the multi-head self-attention layer is illustrated in Fig.~\ref{fig:model}.
	
	Then multi-head self-attention maps the matrix $ \mathbf{H^{A}} $ to query, key, value matrices $ h $ times by different linear projections. 
		\begin{align}
			\mathbf{Q_i}&=\mathbf{H^{A}W_{i}^{Q}}  \\
			\mathbf{K_i}&=\mathbf{H^{A}W_{i}^{K}}  \\
			\mathbf{V_i}&=\mathbf{H^{A}W_{i}^{K}}, i\in 1,\dots,h
		\end{align}
	Where $ \mathbf{W_{i}^{Q}}\in \mathbb{R}^{d\times \frac{d}{h}} $, $ \mathbf{W_{i}^{K}}\in \mathbb{R}^{d\times \frac{d}{h}} $, $ \mathbf{W_{i}^{V}}\in \mathbb{R}^{d\times \frac{d}{h}} $ are the trainable parameters of linear projectors. Each of the projected query matrices $ \mathbf{Q_i} $, key matrices $ \mathbf{K_i} $ and value matrices $ \mathbf{V_i} $ perform the scaled dot-product attention \cite{vaswani2017attention} in parallel. 
		\begin{equation}
			\mathbf{h_{i}}=Attention\left(\mathbf{Q_i},\mathbf{K_i},\mathbf{V_i}\right)
		\end{equation}
	The scaled dot-product attention can be computed as follows:
		\begin{equation}
			Attention\left(\mathbf{Q},\mathbf{K},\mathbf{V}\right)=softmax\left(\frac{\mathbf{Q}\mathbf{K^{T}}}{\sqrt{d}}\right)\mathbf{V}
		\end{equation}
	These attentions are concatenated and projected again to obtain a new representation $ \mathbf{H^{M}} $.
		\begin{equation}
			\mathbf{H^{M}}=concat\left(\mathbf{h_1},\dots,\mathbf{h_{h}}\right)\mathbf{W^{O}}
		\end{equation}
	Where $ \mathbf{W^{O}}\in \mathbb{R}^{d\times d} $ is the trainable parameter of linear projector. where $ \mathbf{H^{M}} $  is the output of the multi-head self-attention layer.
	
	As in the vanilla Transformer \cite{vaswani2017attention}, we add a residual connection and a layer normalization after the multi-head self-attention layer. 
		\begin{equation}
			\mathbf{H^L}=LayerNorm\left ( \mathbf{H}+\mathbf{H^{M}} \right )
		\end{equation}
	Finally, we add a two-layer fully connected feed-forward network after the residual connection and the layer normalization. The output of the attention module can be expressed as,
		\begin{equation}
			[\mathbf{H^I},\mathbf{H^S}]=max\left(0, 	\mathbf{H^{L}W_1+b_1}\right)\mathbf{W_2}+\mathbf{b_2}
		\end{equation}
	where $ \mathbf{H^I}\in \mathbb{R}^{1\times d} $ represents the intent information, $\mathbf{H^S}\in \mathbb{R}^{T\times d} $ represents the slot information for $T$ input tokens, and $ \mathbf{W_{1}} $, $\mathbf{W_{2}}$, $\mathbf{b_{1}}$, and $\mathbf{b_{2}}$ are trainable parameters.
	
	\subsection{Decoder Module}
	The decoder module decodes the intent label $y^I$ and the slot label $y^S$ from $\mathbf{H^I}$ and $\mathbf{H^S}$ via two different fully connected feed-forward networks, respectively. To predict the intent label, we have
		\begin{equation}
			y^{I}=softmax\left(\mathbf{H^I}\mathbf{W^{I}}+\mathbf{b^{I}}\right),
		\end{equation}
	where $ \mathbf{W^{I}} $ and $ \mathbf{b^{I}} $ are trainable parameters. To predict the slot labels $\{y_{i}^{S}\}$, we have
		\begin{equation}
			y_{i}^{S}=softmax\left(\mathbf{h_{i}^{S}}\mathbf{W^{S}+b^{S}}\right),i\in \mathcal{T},
		\end{equation}
	where $\mathbf{h_{i}^{S}}$ is the $i$-th element of $\mathbf{H^{S}}$, and $\mathbf{W^{S}} $ and $ \mathbf{b^{S}} $ are trainable parameters. 
	
	\subsection{Joint training}
	To model two tasks simultaneously, we employ joint optimization to update the parameters of the model. Given the input utterance $ \mathbf{x} $, the conditional probability
	of the understanding result (intent detection and slot filling) is as follows:
		\begin{equation}
			p\left(y^{I},y^{S}\Big|\mathbf{x}\right)=p\left(y^{I}\Big|\mathbf{x}\right)\prod_{n=1}^{N}p\left(y_{n}^{S}\Big|\mathbf{x}\right)
		\end{equation}
	
	When training the model, our loss function can be divided into two parts: intent and slot. The intent loss function can be expressed as follows:
		\begin{equation}
			\mathcal{L}_{ID}=-\sum_{i=1}^{N^{I}}\hat{y}^{i,I}log\left ( y^{i,I} \right )
		\end{equation}
	Similarly, the slot loss function can be expressed as:
		\begin{equation}
			\mathcal{L}_{SF}=-\sum_{j=1}^{T}\sum_{i=1}^{N^{S}}\hat{y}_{j}^{i,S}log\left ( y_{j}^{i,S} \right )
		\end{equation}
	where $ \hat{y}^{i,I} $ and $ \hat{y}_{j}^{i,S} $ represent the target intent label and target slot label separately. In order to jointly train the intent detection and slot filling tasks, the final loss function of two tasks is formulated as:
		\begin{equation}
			\mathcal{L} = \lambda \mathcal{L}_{ID}+(1-\lambda) \mathcal{L}_{SF}
            \label{e:jointLoss}
		\end{equation}
	where the hyper-parameter $ \lambda $ is a mixture weight: $ 0< \lambda < 1 $.
	
	The training objective is to minimize the loss function $ \mathcal{L} $ of two tasks to maximize the conditional probability $ p\left(y^{I},y^{S}|\mathbf{x}\right) $.
	
	\section{Experiments} \label{sec:result}
	\subsection{Datasets}
	We evaluate the proposed FAN framework on two public benchmark datasets, ATIS (Air-line Travel Information System) \cite{tur2010left} and Snips \cite{coucke2018snips}. The ATIS dataset contains audio recordings of people making flight reservations, labeled by 21 intents and 120 slots. The Snips dataset was collected from the Snips personal voice assistant. The statistics of both datasets are summarized in Table~\ref{tab:datasets}.
	\begin{table}
		\centering
		\caption{Statistics of the datasets.}
		\label{tab:datasets}
		\begin{tabular}{lrr}
			\toprule
							   	 & ATIS       & Sinps            \\
			\midrule
			Training set size    & 4478       & 13084            \\
			Development  set size & 500        & 700              \\
			Testing set size     & 893        & 700              \\
			Num Intents          & 21         & 7                \\
			Num Slots            & 120        & 72               \\
			Domain               & air travel & personal assist. \\
			\bottomrule
		\end{tabular}
	\end{table}

	\subsection{Metrics}
	We evaluate FAN under different accuracy and latency performance metrics. Following the conventions of previous work \cite{goo2018slot}, we choose the accuracy for intent detection and the F1 score for slot filling. Furthermore, the sentence-level semantic frame accuracy represents the proportion of utterances that both intent detection and slot filling tasks are predicted correctly. We evaluate the inference latency of different models on multiple platforms, i.e., RTX 3090, Jetson XT2, and Jetson Nano. We feed all samples in the test set one by one into the model and calculate the average latency for predicting one utterance.
	
	\subsection{Training Details}
    \begin{table}
        \centering
        \caption{Training statistics of different FAN models.}
        \label{tab:encoder}
        \begin{tabular}{lrrrr}
            \toprule
            \multirow{2}{*}{Encoder}      & {Transformer} &{Hidden} & \multirow{2}{*}{Parameters (M)} &Training \\
                  & Blocks  &{States $d$} &  & Time (s) \\
            \midrule
            BERT       & 12               & 768                    & 116.6     & 1,456                       \\
            DistilBERT & 6          & 768                   & 59.9        & 857                  \\
            TinyBERT   & 4               & 312                & 15.8       & 485
            \\
            \bottomrule                  
        \end{tabular}
    \end{table}
	We evaluate three FAN  models with different encoders, i.e., BERT, DistilBERT, and TinyBERT, and denote them as BERT-FAN, DistilBERT-FAN, and TinyBERT-FAN, respectively. As shown in the table~\ref{tab:encoder}, BERT has 12 transformer blocks and 768 hidden states, DistilBERT has six transformer blocks and 768 hidden states, and TinyBERT has four transformer blocks and 312 hidden states. The hyper-parameter $ \lambda $ is set to 0.5. The maximum length of utterances is 50. The training batch size is 32. The dropout ratio is  0.1. We use Adam \cite{kingma2014adam} to optimize FAN parameters with a learning rate of 5e-5. Specifically, TinyBERT-FAN only has 15.8M parameters and consumes 485s training time.
	
	\subsection{Benchmarks}
	We compare FAN with the following six benchmarks:
	\begin{itemize}
		\item \textit{Slot-Gated Full Attention}: Goo et al. \cite{goo2018slot} utilized a slot-gated mechanism as a particular gate function in bidirectional long short-term memory (BiLSTM) to improve slot filling using the learned intent context vector. 
		\item \textit{SF-ID Network}: E et al. \cite{niu2019novel} proposed an SF-ID network that consists of an SF subnet and an ID subnet after the BiLSTM encoder. The SF subnet applies intent information to the slot filling task, while the ID subnet uses slot information in the intent detection task. SF-ID network builds a bidirectional connection between intent detection and slot filling to help them promote each other mutually.
		\item \textit{Stack-Propagation}: Qin et al. \cite{qin-etal-2019-stack} adopted a joint BiLSTM-based model with Stack-Propagation, which transmits intent information to slot filling to improve the performance of the slot filling task.
		\item \textit{SlotRefine}: Wu et al. \cite{wu2020slotrefine} proposed a non-autoregressive transformer-based model and designed a two-pass iteration mechanism to handle the problem of the uncoordinated slots and speed up the decoding in slot filling.
		\item \textit{Co-Interactive}: Qin et al. \cite{qin2021co} used BiLSTM as the encoder and proposed a co-interactive transformer to exchange the mutual information of intent and slot. The co-interactive transformer uses two different label attentions for the representation vectors of intents and slots, respectively, and uses two self-attention to exchange information between intent and slots. They concatenated two co-interactive transformers to further enhance the accuracy.
        \item \textit{JointBERT}: Chen et al. \cite{chen2019bert} proposed a BERT-based model with an encoder module and a decoder module. JointBERT connects the representation vector of the first token to a single-layer FFN for intent recognition and connects the representation vector of the remaining tokens to another single-layer FFN for slot filling. The model JointBERT-CRF means that CRF is used for slot decoding. If we remove the attention module from BERT-FAN, i.e., $[\mathbf{H^I},\mathbf{H^S}]=\mathbf{H}$, it coincides with the scheme of JointBERT \cite{chen2019bert}. For better comparison, we further extend JointBERT to the other encoders and denote them as JointDistilBERT and JointTinyBERT in this paper. 
	\end{itemize}

	\subsection{Main Results}
    	\begin{figure}
            \centering
            \begin{subfigure}{.8\columnwidth}
                \centering
                \includegraphics[width=1 \columnwidth]{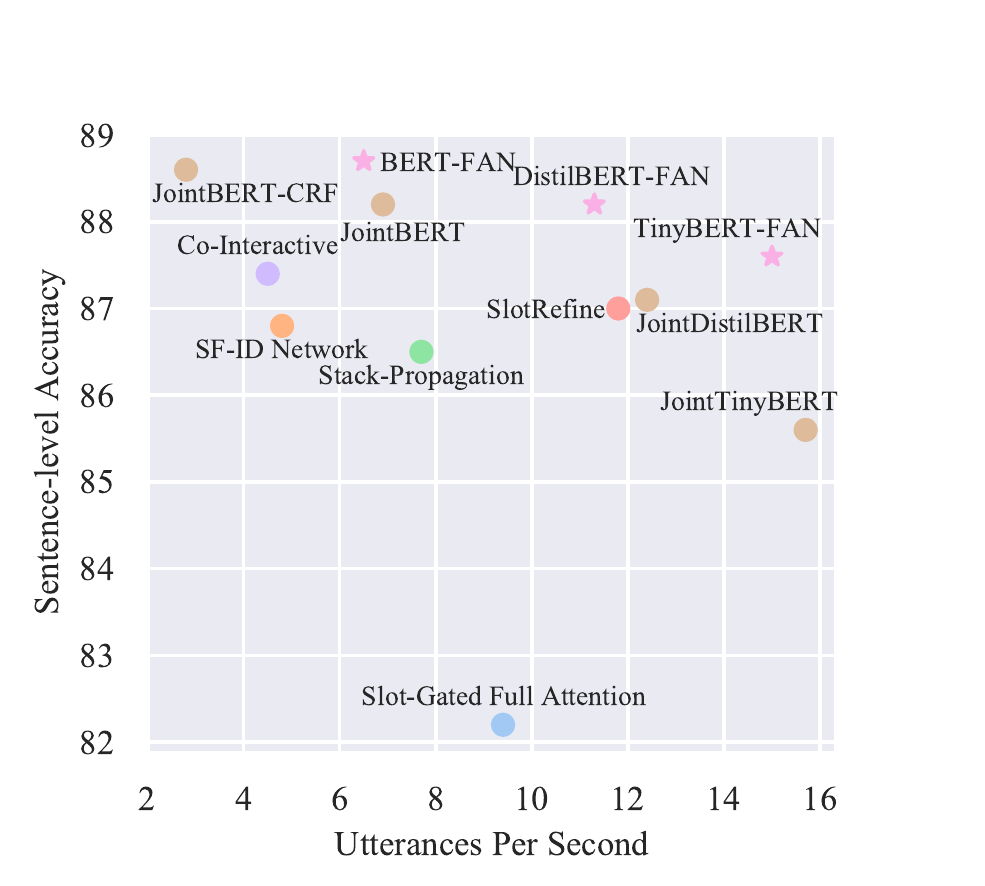}
                \caption{ATIS}
                \label{fig:overall_atis}
            \end{subfigure}
            \begin{subfigure}{.8\columnwidth}
                \centering
                \includegraphics[width=1 \columnwidth]{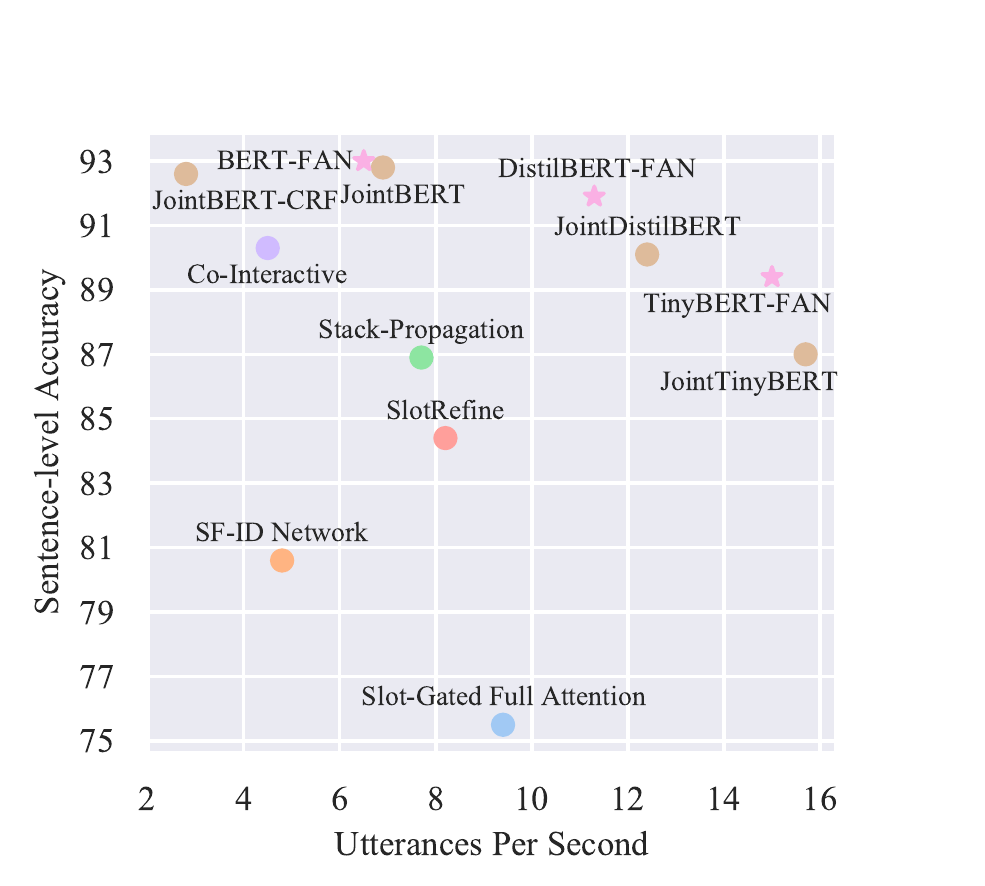}
                \caption{Snips}
                \label{fig:overall_snips}
            \end{subfigure}
        \centering 
        \caption{Inference accuracy vs. speed for different models evaluated on the Jetson Nano platform.}
        \label{fig:overall}
        \end{figure}
    In Fig.~\ref{fig:overall}, we compare the semantic accuracy of FAN to other models with regard to the inference speed. The number of utterances per second for each model is evaluated on the Jetson Nano platform in the 5 Watt mode. FAN delivers more accurate models at every speed level on both ATIS and Snips datasets. The attention module improves the semantic accuracy for different encoders, especially the small-size ones. Compared with jointTinyBERT, TinyBERT-FAN improves the semantic accuracy by more than 2.0\%. TinyBERT-FAN achieves higher accuracy and inferences five more utterances per second than SlotRefine. DistilBERT-FAN balances the inference accuracy and speed.

    \begin{table*}
        \centering
        \caption{Inference accuracy and latency for different models evaluated on the Jetson Nano platform.}
        \label{tab:main}
        \begin{tabular}{lrrrrrrrr}
            \toprule
            \multirow{2}{*}{Model}   					 & \multicolumn{3}{c}{ATIS}    			  & \multicolumn{3}{c}{Snips}  				   & \multirow{2}{*}{Latency (ms)}  & \multirow{2}{*}{Speedup} \\
            & Intent (Acc) & Slot (F1) & Sent (Acc)  & Intent (Acc)   & Slot (F1)   & Sent (Acc)  &                                &                          \\
            \midrule
            Slot-Gated Full Attention \cite{goo2018slot} & 93.6   		& 94.8 		& 82.2        & 97.0   		   & 88.8 		 & 75.5        & 106.9       				    & 3.3x    \\
            SF-ID Network \cite{niu2019novel}            & 97.8  	    & 95.8	    & 86.8        & 97.4   		   & 91.4 		 & 80.6        & 207.2       				    & 1.7x    \\
            Stack-Propagation \cite{qin-etal-2019-stack} & 96.9   		& 95.9 		& 86.5        & 98.0   		   & 94.2 		 & 86.9        & 129.9       				  	& 2.7x    \\
            SlotRefine \cite{wu2020slotrefine}           & 97.1   		& \textbf{96.2} & 87.0    & 97.4  		   & 93.7 		 & 84.4        & 103.4       				  	& 3.4x    \\
            Co-Interactive \cite{qin2021co}		  		 & 97.7   		& 95.9 		& 87.4		  & \textbf{98.8}  & 95.9 		 & 90.3        & 220.3						    & 1.6x    \\
            JointBERT \cite{chen2019bert}                & 97.5   		& 96.1 		& 88.2        & 98.6   		   & 97.0 		 & 92.8        & 145.6       				  	& 2.4x    \\ 
            JointBERT-CRF \cite{chen2019bert}            & \textbf{97.9}  & 96.0    & 88.6 		  & 98.4   	   	   & 96.7  		 & 92.6 	   & 351.0         					& 1.0x   \\
            \midrule
            BERT-FAN       								 & 97.8         & 96.1 		& \textbf{88.7} & 98.3     	   & \textbf{97.1} & \textbf{93.0} & 153.9          			& 2.3x    \\
            DistilBERT-FAN 								 & \textbf{97.9} & 95.9 	& 88.2     	  & 98.0           & 96.5        & 91.9        & 88.2          					& 4.0x    \\
            TinyBERT-FAN   								 & 97.8         & 95.6      & 87.6        & 98.1           & 95.4        & 89.4        & \textbf{66.8} 					& \textbf{5.3x}    \\
            \bottomrule
        \end{tabular}
    \end{table*}
	In Table~\ref{tab:main}, we illustrate the evaluated models' detailed accuracy and latency performance. Based on the same BERT model, BERT-FAN inferences 2.3x faster and achieves comparable or even better performance than the state-of-the-art JointBERT-CRF. It verifies the effectiveness of the proposed attention module in the interaction between intent and slot information. Furthermore, DistillBERT-FAN and TinyBERT-FAN are the only two models whose inference latency is less than 100ms.

	\subsection{Analysis}
	\subsubsection{Ablation Experiments}
    \begin{table*}
        \centering
        \caption{Ablation experiments on the Snips and ATIS datasets.}
        \label{tab:attention}
        \begin{tabular}{lrrrrrr}
            \toprule
            \multirow{2}{*}{Model}        			& \multicolumn{3}{c}{ATIS}    				& \multicolumn{3}{c}{Snips}           		\\
            & Intent (Acc) & Slot (F1)  & Sent (Acc)  	& Intent (Acc) & Slot (F1)  & Sent (Acc)     \\
            \midrule
            TinyBERT-FAN   	  						& 97.8   & \textbf{95.6}  & \textbf{87.6}   & 98.1   & \textbf{95.4}  & \textbf{89.4}   \\
            - without label attention layer 			& 97.6   	   & 95.2  		& 86.9  		& 97.7   	   & 94.9  		& 88.2          \\
            - without multi-head self-attention layer & 97.1   	   & 95.1  		& 85.7  		& 98.1   	   & 94.3  		& 87.0          \\
            - without two-layer FFN  					& \textbf{97.9}  & 95.3  	& 87.1			& \textbf{98.3}  & 94.9  	& 88.6			\\
            \bottomrule
        \end{tabular}
    \end{table*}
    In Table~\ref{tab:attention}, we conduct ablation experiments on each component of the attention module based on TinyBERT-FAN. We remove the label attention layer by feeding the output of the encoder directly into the multi-head self-attention layer, i.e., $\mathbf{H^{A}} = \mathbf{H}$. We remove the multi-head self-attention layer by feeding the output of the label attention layer directly into the decoder, i.e., $\mathbf{H^M} = \mathbf{H^{A}}$. We remove the two-layer fully connected feed-forward network by feeding the output of the residual connection and the layer normalization directly into the decoder module, i.e., $[\mathbf{H^I},\mathbf{H^S}] = \mathbf{H^L}$.
    
    Removing either the label attention layer or the multi-head self-attention layer degrades the accuracy of both intent detection and slot filling. And the latter plays a more critical role than the former, with around a 2\% decrease in the semantic accuracy. Removing the two-layer FFN from TinyBERT-FAN increases the accuracy of intent detection but decreases the slot filling on both ATIS and Snipts datasets. Hence, the FFN layer in the attention module balances intent detection and slot filling to minimize the joint loss function in (\ref{e:jointLoss}). 
    
    \subsubsection{Multi-Head Self-Attention}
     \begin{figure*}
        \centering 
        \includegraphics[width=0.9\textwidth]{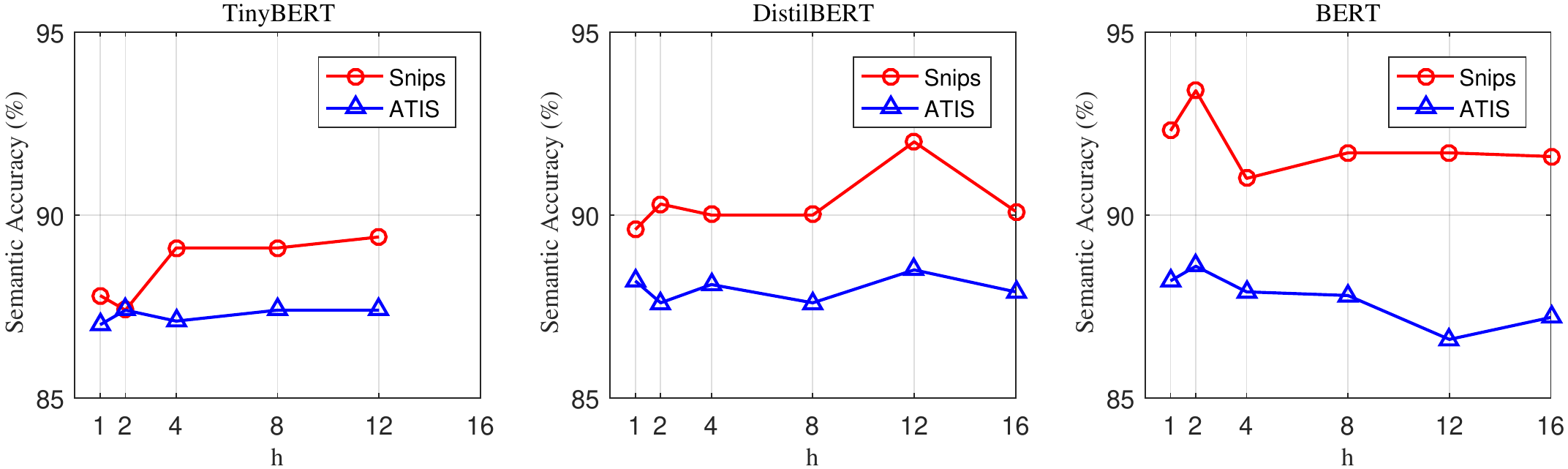}
        \caption{Semantic accuracy of TinyBERT-FAN, DistilBERT-FAN, and BERT-FAN with different number of attention heads, $h$.}
        \label{fig:Head}
    \end{figure*}
    As shown in Fig.~\ref{fig:Head}, we evaluate FAN with different number of attention heads, $h=\{1,2,4,8,12,16\}$. Since TinyBERT has the hidden states $d=312$, which cannot be split equally into sixteen heads, we ignore $h=16$ for TinyBERT-FAN. The optimal $h$ for TinyBERT, DistilBERT, and BERT models are 12, 12, and 2, respectively, on the ATIS and Snips datasets. As illustrated in Table~\ref{tab:attention}, the multi-head self-attention layer plays a key role in the attention module. TinyBERT and DistilBERT tend to choose a greater number of heads than BERT. It explains the observation in Fig.~\ref{fig:overall} that FAN achieves a more significant increase in accuracy when TinyBERT and DistilBERT are used.

	\subsubsection{Information Exchange}
    \begin{table*}
        \centering
        \caption{Comparison between FAN, SF-ID, and Co-interactive on the SNIPS and ATIS datasets.}
        \label{tab:SF-ID}
        \begin{tabular}{lrrrrrrr}
            \toprule
            \multirow{2}{*}{Model}             			& \multicolumn{3}{c}{ATIS}       			& \multicolumn{3}{c}{Snips}         		& \multirow{2}{*}{Latency (ms)}		\\
            & Intent (Acc) & Slot (F1)  & Sent (Acc)  	& Intent (Acc) & Slot (F1)  & Sent (Acc)   	&		  							\\			  
            \midrule
            \multicolumn{1}{l}{JointTinyBERT}       			& 97.1   	   & 94.6  		& 85.5  		& 98.0   	   & 94.4  		& 87.0          & \textbf{63.6}      						\\
            \multicolumn{1}{l}{TinyBERT-SF-ID} 			& 97.6   	   & 95.3       & 87.1          & 97.8   	   & 94.9  		& 88.2 			& 285.8       						\\
            \multicolumn{1}{l}{TinyBERT-Co-interactive} & 97.4         & \textbf{95.6}  & 87.5      & 98.0   	   & 95.2  		& 89.1 			& 310.8       						\\
            
            \midrule
            TinyBERT-FAN            & \textbf{97.8}   & \textbf{95.6}  & \textbf{87.6}  		& \textbf{98.1}   & \textbf{95.4}  & \textbf{89.4}             & 66.8       						\\
            \bottomrule
        \end{tabular}
    \end{table*}
	In Table~\ref{tab:SF-ID}, we compare FAN with SF-ID and Co-interactive based on the TinyBERT encoder. All three schemes enhance the accuracy by exchanging information between intent and slot and are independent of the encoder module. For a fair comparison, we replace BiLSTM in SF-ID and Co-interactive by TinyBERT and keep the remaining parts unchanged, namely TinyBERT-SF-ID and TinyBERT-Co-interactive. As shown in Table~\ref{tab:SF-ID}, TinyBERT-FAN achieves comparable or even better performance than TinyBERT-SF-ID and TinyBERT-Co-interactive but requires significantly less latency, close to JointTinyBERT. Here, JointTinyBERT is a baseline with no additional layer between the encoder and the decoder. Hence, FAN is a lightweight and efficient network suitable for joint intent detection and slot filling on edge devices.
	
	When we look at the detailed network structure, FAN use one label attention and one multi-head self-attention. Co-interactive concatenates two transformers, and each transformer uses two label attentions and two self-attentions for intent and slot, respectively. SF-ID divides the interaction between intent and slot into two different steps. Both SF-ID and Co-interactive use CRF for the slot decoding, which greatly slows the inference speed.

	\subsubsection{Error Analysis}
	
	\begin{figure}
		\centering 
		\includegraphics{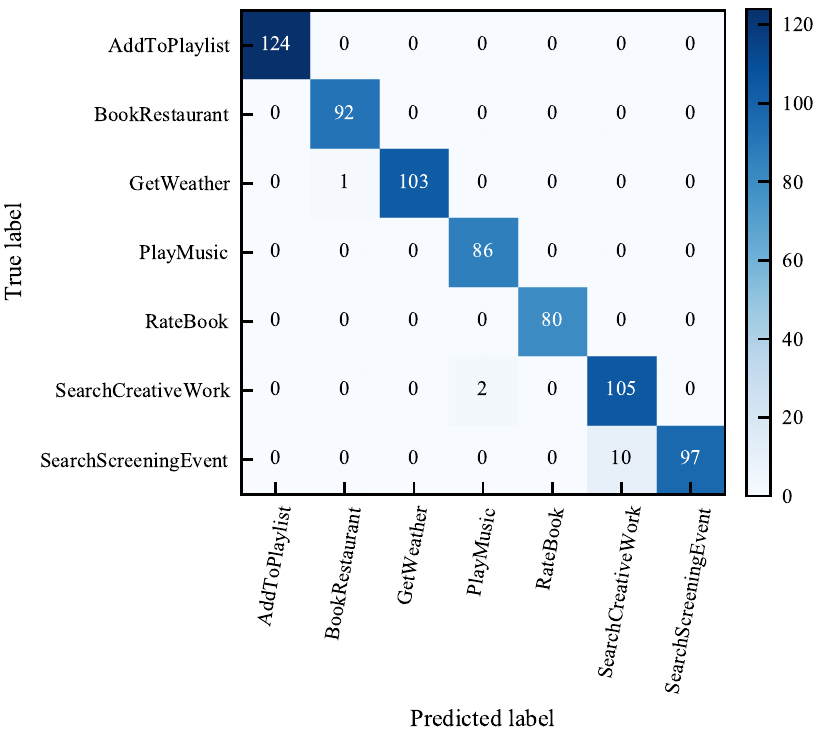}
		\caption{Intent confusion matrix of DistilBERT-FAN on the Snips dataset}
		\label{fig:intent_cm}
	\end{figure}

	\begin{table*}
		\centering
		\caption{Three examples for error analysis}
		\label{tab:error}
		\resizebox{\linewidth}{!}{
		\begin{tabular}{c|lllllll}
			\toprule
			Utterance       	 & Play & the    & album           & journeyman         & & &        	\\ 
			Truth Slot Label     & O    & O		 & B-object\_type  & B-object\_type 	& & &			\\ 
			Predicted Slot Label & O    & O      & B-music\_item   & B-album            & & &    		\\
			Truth Intent Label	 & Search Creative Work 							&&&	& & & 			\\
			Predicted Intent Label & Play Music										&&&	& & &			\\
			\midrule
			Utterance      		 & Play & the            & new            & noise          & theology       & e 				& p              \\ 
			Truth Slot Label     & O    & B-object\_name & I-object\_name & I-object\_name & I-object\_name & I-object\_name & I-object\_name \\ 
			Predicted Slot Label & O    & B-album        & I-album        & I-album        & I-album        & I-album        & I-album        \\
			Truth Intent Label	 & Search Creative Work 							&&&	& & & 			\\
			Predicted Intent Label & Play Music										&&&	& & &			\\
			\midrule
			Utterance      		 & In 		  & one 		& hour 		  & find & king 		  & of 			   & hearts                   \\ 
			Truth Slot Label     & B-timeRange & I-timeRange & I-timeRange & O 	 & B-movie\_name  & I-movie\_name  & I-movie\_name            \\
			Predicted Slot Label & B-timeRange & I-timeRange & I-timeRange & O 	 & B-object\_name & I-object\_name & I-object\_name           \\
			Truth Intent Label	 & Search Screening Event							&&&	& & & 			\\
			Predicted Intent Label & Search Creative Work							&&&	& & &			\\
			\bottomrule
		\end{tabular}}
	\end{table*}

In Fig.~\ref{fig:intent_cm}, we plot the confusion matrix of intent prediction on the test dataset. Only 13 out of 700 utterances are misclassified, among which seven utterances with the intent ``Search Screening Event'' are mistakenly predicted to be ``Search Creative Work''. We demonstrate three error cases in Table~\ref{tab:error}. First, the utterance ``Play the album journeyman'' actually means to search and play "journeyman" and is labeled ``Search Creative Work''. However, since ``journeyman'' is an album, the model predicts the intent of ``Play Music'' based on the keyword ``play'' \cite{wu-etal-2021-label}. Considering the second utterance ``Play the new noise theology ep'', the slot label of ``the new noise theology ep'' is ``object\_name''. However, ``the new noise theology '' is the name of the album. TinyBERT-FAN learns this information with pre-trained knowledge from Wikipedia and predicts it as ``album''. ``king of hearts'' in the third utterance is the name of a comedy movie. However, TinyBERT-FAN fails to recognize the movie and mistakenly predicts it as ``object\_name''.
	
	\subsection{Edge Deployment}

	\begin{table*}[]
		\centering
		\caption{Inference speed on different devices}
		\label{tab:edge}
		\begin{tabular}{ccrrrrrrrr}
			\toprule
			\multirow{2}{*}{Devices}     & \multirow{2}{*}{Mode}  & \multicolumn{2}{c}{JointBERT-CRF} & \multicolumn{2}{c}{BERT-FAN}  & \multicolumn{2}{c}{DistilBERT-FAN} & \multicolumn{2}{c}{TinyBERT-FAN} \\
			&                       	 						  & Latency (ms)         & Speedup    & Latency (ms)     & Speedup     & Latency (ms)    & Speedup          & Latency (ms)      & Speedup      \\
			\midrule
			RTX 3090					 &			/			  & 18.5			     & 1.0x		  & 8.1				 & 2.3x		   & 4.7   			 & 3.9x				& 3.9				& 4.7x		   \\
			Jetson XT2                   & MAX-P ARM              & 114.9                & 1.0x       & 63.1             & 1.9x        & 38.3            & 3.0x             & 26.8              & 4.3x         \\
			\multirow{2}{*}{Jetson Nano} & MAX-N                  & 223.9                & 1.0x       & 115.7            & 1.9x        & 69.0            & 3.2x             & 40.4              & 5.5x         \\
										 & 5W                     & 351.0                & 1.0x       & 153.9            & 2.3x        & 88.2            & 4.0x             & 66.8              & 5.3x  		\\      
			\bottomrule  
		\end{tabular}
	\end{table*}

	In Table~\ref{tab:edge}, we deploy and evaluate the inference latency of FAN on the desktop platform NVIDIA GeForce RTX 3090 and popular edge platforms, i.e., Jetson XT2 and Jetson Nano. We choose the default MAX-P ARM mode for Jetson XT2 and test Jetson Nano in both MAX-N and 5W modes. The 5W mode limits the Jetson Nano's power to 5w and limits the number of CPU cores and CPU and GPU frequencies. We include JointBERT-CRF as a baseline. From Table~\ref{tab:edge}, TinyBERT-FAN achieves 4.3x speedup on Jetson XT2 and 5.3x speedup on Jetson Nano in the 5W mode. The inference accuracy of these deployed models is consistent with the performance in Table~\ref{tab:main}. Hence, both DistilBERT-FAN and TinyBERT-FAN significantly reduce the inference latency to less than 100ms and are suitable for deploying on different edge devices.

	\section{Conclusion} \label{sec:conclusion}
	In this paper, we propose a fast attention network based on pre-trained language models for joint intent detection and slot filling tasks. The experimental results show that FAN delivers more accurate models at every speed level on two public datasets. The proposed attention module improves the semantic accuracy by more than 2.0\% when TinyBERT is the encoder. Moreover, we deploy FAN on popular edge platforms, which inferences fifteen utterances per second on the Jecson Nano platform. We conclude that FAN has experimentally proven its value in industrial practice for deployment at the edge. 
    
    For future work, we plan to employ an integer-only encoder, i.e., I-BERT \cite{kim2021bert}, to further reduce the inference latency. In addition, we intend to incorporate external knowledge via graph neural networks to further enhance accuracy.

\bibliographystyle{IEEEtran}
\bibliography{IEEEabrv,Reference}

\begin{thebibliography}{10}
\providecommand{\url}[1]{#1}
\csname url@samestyle\endcsname
\providecommand{\newblock}{\relax}
\providecommand{\bibinfo}[2]{#2}
\providecommand{\BIBentrySTDinterwordspacing}{\spaceskip=0pt\relax}
\providecommand{\BIBentryALTinterwordstretchfactor}{4}
\providecommand{\BIBentryALTinterwordspacing}{\spaceskip=\fontdimen2\font plus
\BIBentryALTinterwordstretchfactor\fontdimen3\font minus
  \fontdimen4\font\relax}
\providecommand{\BIBforeignlanguage}[2]{{%
\expandafter\ifx\csname l@#1\endcsname\relax
\typeout{** WARNING: IEEEtran.bst: No hyphenation pattern has been}%
\typeout{** loaded for the language `#1'. Using the pattern for}%
\typeout{** the default language instead.}%
\else
\language=\csname l@#1\endcsname
\fi
#2}}
\providecommand{\BIBdecl}{\relax}
\BIBdecl

\bibitem{bertino2020artificial}
E.~Bertino and S.~Banerjee, ``Artificial intelligence at the edge,'' 2020,
  arXiv:2012.05410.

\bibitem{xu2020cha}
L.~Xu, A.~Iyengar, and W.~Shi, ``{CHA}: A caching framework for home-based
  voice assistant systems,'' in \emph{Proc. IEEE/ACM Symp. Edge Comput.}, 2020,
  pp. 293--306.

\bibitem{pandelea2021emotion}
V.~Pandelea, E.~Ragusa, T.~Apicella, P.~Gastaldo, and E.~Cambria, ``Emotion
  recognition on edge devices: Training and deployment,'' \emph{Sensors},
  vol.~21, no.~13, p. 4496, 2021.

\bibitem{young2013pomdp}
S.~Young, M.~Gašić, B.~Thomson, and J.~D. Williams, ``Pomdp-based statistical
  spoken dialog systems: A review,'' \emph{Proceedings of the IEEE}, vol. 101,
  no.~5, pp. 1160--1179, 2013.

\bibitem{tur2011spoken}
G.~Tur and R.~De~Mori, \emph{Spoken language understanding: Systems for
  extracting semantic information from speech}.\hskip 1em plus 0.5em minus
  0.4em\relax John Wiley \& Sons, 2011.

\bibitem{jeong2008triangular}
M.~Jeong and G.~G. Lee, ``Triangular-chain conditional random fields,''
  \emph{{IEEE/ACM} Trans. Audio, Speech, Language Process.}, vol.~16, no.~7,
  pp. 1287--1302, 2008.

\bibitem{haffner2003optimizing}
P.~Haffner, G.~Tur, and J.~H. Wright, ``Optimizing {SVM}s for complex call
  classification,'' in \emph{Proc. IEEE Int. Conf. Acoust., Speech, Signal
  Process}, vol.~1, 2003, pp. I--I.

\bibitem{yao2014spoken}
K.~Yao, B.~Peng, Y.~Zhang, D.~Yu, G.~Zweig, and Y.~Shi, ``Spoken language
  understanding using long short-term memory neural networks,'' in \emph{Proc.
  IEEE Spoken Lang. Technol. Workshop}, 2014, pp. 189--194.

\bibitem{raymond2007generative}
C.~Raymond and G.~Riccardi, ``Generative and discriminative algorithms for
  spoken language understanding,'' in \emph{Proc. Interspeech}, 2007.

\bibitem{chen2016syntax}
Y.-N. Chen, D.~Hakanni-T{\"u}r, G.~Tur, A.~Celikyilmaz, J.~Guo, and L.~Deng,
  ``Syntax or semantics? knowledge-guided joint semantic frame parsing,'' in
  \emph{Proc. IEEE Spoken Lang. Technol. Workshop}, 2016, pp. 348--355.

\bibitem{zhang2016joint}
X.~Zhang and H.~Wang, ``A joint model of intent determination and slot filling
  for spoken language understanding.'' in \emph{Proc. Int. Joint Conf. Artif.
  Intell.}, vol.~16, 2016, pp. 2993--2999.

\bibitem{hakkani2016multi}
D.~Hakkani-T{\"u}r, G.~T{\"u}r, A.~Celikyilmaz, Y.-N. Chen, J.~Gao, L.~Deng,
  and Y.-Y. Wang, ``Multi-domain joint semantic frame parsing using
  bi-directional {RNN-LSTM},'' in \emph{Proc. Interspeech}, 2016, pp. 715--719.

\bibitem{liu2016attention}
B.~Liu and I.~Lane, ``Attention-based recurrent neural network models for joint
  intent detection and slot filling,'' 2016, arXiv:1609.01454.

\bibitem{goo2018slot}
C.-W. Goo, G.~Gao, Y.-K. Hsu, C.-L. Huo, T.-C. Chen, K.-W. Hsu, and Y.-N. Chen,
  ``Slot-gated modeling for joint slot filling and intent prediction,'' in
  \emph{Proc. NAACL}, New Orleans, Louisiana, USA, Jun. 2018, pp. 753--757.

\bibitem{qin-etal-2019-stack}
L.~Qin, W.~Che, Y.~Li, H.~Wen, and T.~Liu, ``A stack-propagation framework with
  token-level intent detection for spoken language understanding,'' in
  \emph{Proc. EMNLP-IJCNLP}, Hong Kong, China, Nov. 2019, pp. 2078--2087.

\bibitem{qin2021co}
L.~Qin, T.~Liu, W.~Che, B.~Kang, S.~Zhao, and T.~Liu, ``A co-interactive
  transformer for joint slot filling and intent detection,'' in \emph{Proc.
  IEEE Int. Conf. Acoust., Speech, Signal Process}, 2021, pp. 8193--8197.

\bibitem{devlin-etal-2019-bert}
J.~Devlin, M.-W. Chang, K.~Lee, and K.~Toutanova, ``{BERT}: Pre-training of
  deep bidirectional transformers for language understanding,'' in \emph{Proc.
  NAACL}, Minneapolis, Minnesota, USA, 2019, pp. 4171--4186.

\bibitem{chen2019bert}
Q.~Chen, Z.~Zhuo, and W.~Wang, ``{BERT} for joint intent classification and
  slot filling,'' 2019, arXiv:1902.10909.

\bibitem{zhang2019joint}
Z.~Zhang, Z.~Zhang, H.~Chen, and Z.~Zhang, ``A joint learning framework with
  {BERT} for spoken language understanding,'' \emph{IEEE Access}, vol.~7, pp.
  168\,849--168\,858, 2019.

\bibitem{wu2020slotrefine}
\BIBentryALTinterwordspacing
D.~Wu, L.~Ding, F.~Lu, and J.~Xie, ``{SlotRefine}: A fast non-autoregressive
  model for joint intent detection and slot filling,'' in \emph{Proc. EMNLP},
  Nov. 2020, p. 1932–1937. [Online]. Available:
  \url{https://aclanthology.org/2020.emnlp-main.152}
\BIBentrySTDinterwordspacing

\bibitem{khattab2020colbert}
\BIBentryALTinterwordspacing
O.~Khattab and M.~Zaharia, ``Colbert: Efficient and effective passage search
  via contextualized late interaction over bert,'' in \emph{Proc. 43rd Int. ACM
  SIGIR Conf. Res. Develop. Inf. Retrieval}, New York, NY, USA, 2020, p.
  39–48. [Online]. Available: \url{https://doi.org/10.1145/3397271.3401075}
\BIBentrySTDinterwordspacing

\bibitem{liu2021bringing}
\BIBentryALTinterwordspacing
D.~Liu, H.~Kong, X.~Luo, W.~Liu, and R.~Subramaniam, ``Bringing ai to edge:
  From deep learning’s perspective,'' \emph{Neurocomputing}, vol. 485, pp.
  297--320, 2022. [Online]. Available:
  \url{https://www.sciencedirect.com/science/article/pii/S0925231221016428}
\BIBentrySTDinterwordspacing

\bibitem{sanh2019distilbert}
V.~Sanh, L.~Debut, J.~Chaumond, and T.~Wolf, ``{DistilBERT}, a distilled
  version of {BERT}: smaller, faster, cheaper and lighter,'' 2019,
  arXiv:1910.01108.

\bibitem{jiao2020tinybert}
\BIBentryALTinterwordspacing
X.~Jiao, Y.~Yin, L.~Shang, X.~Jiang, X.~Chen, L.~Li, F.~Wang, and Q.~Liu,
  ``{T}iny{BERT}: Distilling {BERT} for natural language understanding,'' in
  \emph{Proc. EMNLP}, Nov. 2020. [Online]. Available:
  \url{https://aclanthology.org/2020.findings-emnlp.372}
\BIBentrySTDinterwordspacing

\bibitem{liu2020fastbert}
\BIBentryALTinterwordspacing
W.~Liu, P.~Zhou, Z.~Wang, Z.~Zhao, H.~Deng, and Q.~Ju, ``{F}ast{BERT}: a
  self-distilling {BERT} with adaptive inference time,'' in \emph{Proc. ACL},
  Jul. 2020, pp. 6035--6044. [Online]. Available:
  \url{https://aclanthology.org/2020.acl-main.537}
\BIBentrySTDinterwordspacing

\bibitem{sun2020mobilebert}
\BIBentryALTinterwordspacing
Z.~Sun, H.~Yu, X.~Song, R.~Liu, Y.~Yang, and D.~Zhou, ``{M}obile{BERT}: a
  compact task-agnostic {BERT} for resource-limited devices,'' in \emph{Proc.
  ACL}, Jul. 2020, pp. 2158--2170. [Online]. Available:
  \url{https://aclanthology.org/2020.acl-main.195}
\BIBentrySTDinterwordspacing

\bibitem{kim2021bert}
S.~Kim, A.~Gholami, Z.~Yao, M.~W. Mahoney, and K.~Keutzer, ``I-bert:
  Integer-only bert quantization,'' in \emph{Proc. Int. Conf. Mach. Learn},
  2021, pp. 5506--5518.

\bibitem{wang2020structured}
\BIBentryALTinterwordspacing
Z.~Wang, J.~Wohlwend, and T.~Lei, ``Structured pruning of large language
  models,'' in \emph{Proc. EMNLP}, Nov. 2020, pp. 6151--6162. [Online].
  Available: \url{https://aclanthology.org/2020.emnlp-main.496}
\BIBentrySTDinterwordspacing

\bibitem{li2018self}
\BIBentryALTinterwordspacing
C.~Li, L.~Li, and J.~Qi, ``A self-attentive model with gate mechanism for
  spoken language understanding,'' in \emph{Proc. EMNLP}, Brussels, Belgium,
  Oct.-Nov. 2018, pp. 3824--3833. [Online]. Available:
  \url{https://aclanthology.org/D18-1417}
\BIBentrySTDinterwordspacing

\bibitem{niu2019novel}
\BIBentryALTinterwordspacing
H.~E, P.~Niu, Z.~Chen, and M.~Song, ``A novel bi-directional interrelated model
  for joint intent detection and slot filling,'' in \emph{Proc. ACL}, Florence,
  Italy, Jul. 2019, pp. 5467--5471. [Online]. Available:
  \url{https://aclanthology.org/P19-1544}
\BIBentrySTDinterwordspacing

\bibitem{wei2021joint}
P.~Wei, B.~Zeng, and W.~Liao, ``Joint intent detection and slot filling with
  wheel-graph attention networks,'' \emph{Journal of Intelligent \& Fuzzy
  Systems}, no. Preprint, pp. 1--12, 2021.

\bibitem{yang2019xlnet}
Z.~Yang, Z.~Dai, Y.~Yang, J.~Carbonell, R.~R. Salakhutdinov, and Q.~V. Le,
  ``{XLNet}: Generalized autoregressive pretraining for language
  understanding,'' \emph{Proc. Neural Inf. Process. Syst.}, vol.~32, 2019.

\bibitem{liu2019roberta}
Y.~Liu, M.~Ott, N.~Goyal, J.~Du, M.~Joshi, D.~Chen, O.~Levy, M.~Lewis,
  L.~Zettlemoyer, and V.~Stoyanov, ``{RoBERTa}: A robustly optimized bert
  pretraining approach,'' 2019, arXiv:1907.11692.

\bibitem{lan2019albert}
Z.~Lan, M.~Chen, S.~Goodman, K.~Gimpel, P.~Sharma, and R.~Soricut, ``{ALBERT}:
  A lite bert for self-supervised learning of language representations,'' 2019,
  arXiv:1909.11942.

\bibitem{wang2018glue}
\BIBentryALTinterwordspacing
A.~Wang, A.~Singh, J.~Michael, F.~Hill, O.~Levy, and S.~Bowman, ``{GLUE}: A
  multi-task benchmark and analysis platform for natural language
  understanding,'' in \emph{Proc. EMNLP}, Brussels, Belgium, Nov. 2018, pp.
  353--355. [Online]. Available: \url{https://aclanthology.org/W18-5446}
\BIBentrySTDinterwordspacing

\bibitem{ramshaw1999text}
L.~A. Ramshaw and M.~P. Marcus, ``Text chunking using transformation-based
  learning,'' in \emph{Natural language processing using very large
  corpora}.\hskip 1em plus 0.5em minus 0.4em\relax Springer, 1999, pp.
  157--176.

\bibitem{coucke2018snips}
A.~Coucke, A.~Saade, A.~Ball, T.~Bluche, A.~Caulier, D.~Leroy, C.~Doumouro,
  T.~Gisselbrecht, F.~Caltagirone, T.~Lavril \emph{et~al.}, ``Snips voice
  platform: an embedded spoken language understanding system for
  private-by-design voice interfaces,'' 2018, arXiv:1805.10190.

\bibitem{wu2016google}
Y.~Wu, M.~Schuster, Z.~Chen, Q.~V. Le, M.~Norouzi, W.~Macherey, M.~Krikun,
  Y.~Cao, Q.~Gao, K.~Macherey \emph{et~al.}, ``Google's neural machine
  translation system: Bridging the gap between human and machine translation,''
  2016, arXiv:1609.08144.

\bibitem{he2016deep}
K.~He, X.~Zhang, S.~Ren, and J.~Sun, ``Deep residual learning for image
  recognition,'' in \emph{Proc. IEEE Comput. Soc. Conf. Comput. Vis. Pattern
  Recognit}, 2016, pp. 770--778.

\bibitem{ba2016layer}
J.~L. Ba, J.~R. Kiros, and G.~E. Hinton, ``Layer normalization,'' 2016,
  arXiv:1607.06450.

\bibitem{vaswani2017attention}
A.~Vaswani, N.~Shazeer, N.~Parmar, J.~Uszkoreit, L.~Jones, A.~N. Gomez,
  {\L}.~Kaiser, and I.~Polosukhin, ``Attention is all you need,'' in
  \emph{Proc. Neural Inf. Process. Syst.}, 2017, pp. 6000--6010.

\bibitem{tur2010left}
G.~Tur, D.~Hakkani-T{\"u}r, and L.~Heck, ``What is left to be understood in
  {ATIS}?'' in \emph{Proc. IEEE Spoken Lang. Technol. Workshop}, 2010, pp.
  19--24.

\bibitem{kingma2014adam}
D.~P. Kingma and J.~Ba, ``Adam: A method for stochastic optimization,'' in
  \emph{Proc. Int. Conf. Learn. Representations}, 2015.

\bibitem{wu-etal-2021-label}
\BIBentryALTinterwordspacing
T.-W. Wu, R.~Su, and B.~Juang, ``A label-aware {BERT} attention network for
  zero-shot multi-intent detection in spoken language understanding,'' in
  \emph{Proc. EMNLP}, Online and Punta Cana, Dominican Republic, Nov. 2021, pp.
  4884--4896. [Online]. Available:
  \url{https://aclanthology.org/2021.emnlp-main.399}
\BIBentrySTDinterwordspacing

\end{thebibliography}
\end{document}